\documentclass{edm_article}

\usepackage[protrusion]{microtype} 

\usepackage{multirow} 
\usepackage[table, dvipsnames]{xcolor}
\newcommand{\g}[1]{\textcolor{gray}{#1}}

\usepackage{enumitem} 
\setitemize{itemsep=0pt,topsep=0pt}
\setenumerate{itemsep=0pt,topsep=0pt}

\usepackage[breaklinks,colorlinks,linkcolor=blue,urlcolor=blue,citecolor=blue]{hyperref} 
\usepackage[nameinlink]{cleveref}

\makeatletter
\g@addto@macro{\email}{\normalsize}
\makeatother

\begin{document}

\title{Evaluating Algorithmic Bias in Models for Predicting Academic Performance of Filipino Students}


\numberofauthors{3}
\author{
\alignauthor
Valdemar Švábenský \\
       \affaddr{University of Pennsylvania \& Kyushu University}\\
       \email{valdemar.research@gmail.com}
\alignauthor
Mélina Verger \\
       \affaddr{Sorbonne University, LIP6}\\
       \email{melina.verger@lip6.fr}
\alignauthor
Maria Mercedes T. Rodrigo \\
       \affaddr{Ateneo de Manila University}\\
       \email{mrodrigo@ateneo.edu}
\and  
\alignauthor
Clarence James G. Monterozo \\
       \affaddr{Ateneo de Manila University}\\
       \email{jamesmonterozo@gmail.com}
\alignauthor
Ryan S. Baker\\
       \affaddr{University of Pennsylvania}\\
       \email{ryanshaunbaker@gmail.com}
\alignauthor
Miguel Zenon Nicanor Lerias Saavedra \\
       \affaddr{Ateneo de Manila University}\\
       \email{msaavedra@ateneo.edu}
\and  
\alignauthor
Sébastien Lallé \\
       \affaddr{Sorbonne University, LIP6}\\
       \email{sebastien.lalle@lip6.fr}
\alignauthor
Atsushi Shimada \\
       \affaddr{Kyushu University}\\
       \email{atsushi@ait.kyushu-u.ac.jp}
}

\toappear{\scriptsize V. Švábenský, M. Verger, M. Rodrigo, C. Monterozo, R. Baker, M. Saavedra, S. Lallé, and A. Shimada. Evaluating Algorithmic Bias in Models for Predicting Academic Performance of Filipino Students. In \textit{Proceedings of the 17th International Conference on Educational Data Mining}, pages 744--751, Atlanta, GA, USA, July 2024. Editors: B. Paaßen and C. D. Epp. International Educational Data Mining Society.\\

© 2024 Copyright is held by the author(s). This work is distributed under the Creative Commons Attribution NonCommercial NoDerivatives 4.0 International (CC BY-NC-ND 4.0) license.
\\\url{https://doi.org/10.5281/zenodo.12729936}
}

\maketitle

\begin{abstract}
Algorithmic bias is a major issue in machine learning models in educational contexts. However, it has not yet been studied thoroughly in Asian learning contexts, and only limited work has considered algorithmic bias based on regional (sub-national) background. As a step towards addressing this gap, this paper examines the population of 5,986 students at a large university in the Philippines, investigating algorithmic bias based on students' regional background. The university used the Canvas learning management system (LMS) in its online courses across a broad range of domains. Over the period of three semesters, we collected 48.7 million log records of the students' activity in Canvas. We used these logs to train binary classification models that predict student grades from the LMS activity. The best-performing model reached AUC of 0.75 and weighted F1-score of 0.79. Subsequently, we examined the data for bias based on students' region. Evaluation using three metrics: AUC, weighted F1-score, and MADD showed consistent results across all demographic groups. Thus, no unfairness was observed against a particular student group in the grade predictions.
\end{abstract}


\section{Introduction}
\label{sec_intro}

\textit{Algorithmic bias} in education can be defined in a wide range of ways~\cite{baker2021algorithmic, anderson2019assessing}. In this paper, it denotes \textit{group unfairness}~\cite{barocas2019}, i.e., situations when the performance metrics for a machine learning model substantially differ across groups of students. Here, \textit{group} is defined by a set of immutable characteristics, for example, the student's native language or hometown.

Algorithmic bias can arise in any stage of the educational data mining (EDM) pipeline~\cite{baker2021algorithmic, kizilcec2022algorithmic}.
Subsequently, it can lead to discriminatory treatment or even harm to certain groups of students, negatively affecting their learning~\cite{hu2020towards}. This often comes with severe social, ethical, or legal implications~\cite{kizilcec2022algorithmic}. Therefore, it is essential to study algorithmic bias in order to inform decision-makers about how to possibly mitigate it.

\subsection{Background}

Paquette et al.~\cite{paquette2020} explored how EDM studies employed demographic data, concluding that the field needs more research to validate model performance on various student subpopulations. Baker and Hawn~\cite{baker2021algorithmic}
urged the community to conduct more studies across contexts, indicating that there are two key areas where more work is needed.

First, almost all published research focuses on learners in the United States~\cite{baker2021algorithmic}. Populations in other countries were studied rarely.
As part of addressing this literature gap, our paper studies these issues in relation to learners in the Philippines. Since bias may have different characteristic manifestations in different contexts, it is important to search for ``evidence for algorithmic bias impacting groups that have not been previously studied''~\cite{baker2021algorithmic}. As a result, considering a perspective from another country contributes towards global understanding of the complex issue of algorithmic bias.

Second, previous work often defined groups of learners using distinctions that are too coarse~\cite{baker2021algorithmic}. For example, all Asian-Americans in the published research on algorithmic bias in education were usually treated as the same group, even though Asian countries -- or even regions within a single country -- have numerous cultural differences. The second unique aspect of this paper is investigating fine-grained demographic categories within the geographic regions of the Philippines. This is a response to a call for researchers to explore differences across finely-differentiated groups~\cite{baker2020culture}.

\subsection{Motivation and Local Context}
\label{subsec:motivation}

Studies conducted worldwide during the COVID-19 pandemic revealed that learning losses were larger for lower-income countries~\cite{Wu2022geographic}. In the Philippines, learners from resource-scarce backgrounds encountered greater challenges compared to those with robust access to digital resources~\cite{barrot2021students}. Students from less affluent families grappled with the scarcity of computers, mobile devices, and reliable Internet connection, which greatly limited their ability to fully engage in online learning~\cite{cleofas2021demographic}. Students from rural and low-income areas also reported lower online learner readiness~\cite{Clemen2021Education}.

The Philippines, with its diverse landscape across the individual regions, illustrates this digital divide. Access to the Internet and cellular services within the country varies significantly, depending on socioeconomics and regional geography~\cite{sy2021mapping}. Urban centers generally enjoy high-speed Internet, while rural areas possess limited digital resources. In areas like Metro Manila, Central Luzon, and Cebu, both wealth and Internet speeds are higher.
In contrast, regions such as Palawan, Eastern Visayas, and Northern Mindanao grapple with heightened poverty rates and slower Internet speeds.

\subsection{Research Problem Statement}
\label{subsec:problem-statement}

Our study focuses on students from a Filipino university during 2021--2022. At that time, the Philippines remained under COVID-19 lockdown measures, compelling universities to pivot to remote learning. Students were prohibited from coming on campus and had to complete their academic requirements from home, in different regions of the country. To facilitate online learning, stable Internet connection and technological resources are crucial. However, given the considerable disparities throughout the Philippines, as well as the cultural differences of different regions, we hypothesize that students' academic achievements during this period were significantly influenced by their geographical locations.

This paper explores the following research question: \textit{When building predictive models of academic performance of Filipino students in online courses, do the models exhibit identifiable algorithmic bias towards a certain demographic?} Here, the demographic groups are defined based on the location from which the students accessed the online courses. We assume that the location is also highly indicative of the students' regional background, not only their place of residence. And, since different geographic locations have different properties and constraints on distance learning, we want to see whether failing to account for differences between students in different locations introduces bias in the model predictions.

\section{Related Work}

Since our research focuses on two topics: prediction of academic performance and bias, we review papers in these areas.

\subsection{Student Performance Prediction}

We study predicting performance of students as it is beneficial for teaching faculty and university administrators~\cite{anderson2019assessing}. Instructors can proactively offer assistance and support to students identified as being at risk of poor performance. Administrators can use the prediction results to guide the retention of students. Recently, related work focused on explainable prediction models~\cite{Hoq2023edm, Swamy2022edm}, which would better support the needs of these stakeholders.

Hellas et al.~\cite{Hellas2018} published an extensive review of student performance prediction. They pointed out that the values being predicted can range from \textit{long-term}, such as college graduation~\cite{aulck2019mining}, through \textit{medium-term}, such as course grades~\cite{Ong2022edm}, to \textit{short-term}, such as assignment performance~\cite{hicks2022}. Data sources that provide features for prediction include course data, student demographics, and psychometric information such as motivation. Our work employs course data for prediction and student demographics for bias evaluation. Commonly used classification models in EDM include Decision Tree, Random Forest, K-Nearest Neighbors, eXtreme Gradient Boosting, and Logistic Regression~\cite{Korkmaz}. For the sake of comparison, we use these models as well.

\subsection{Algorithmic Bias in Prediction Models}

Although most EDM research did not address the issue of algorithmic bias~\cite{paquette2020}, several studies have been conducted over the past decade. To prevent bias, researchers argue against using demographic variables as predictors~\cite{Baker2023, Cohausz2023}. In a similar vein, the output of a predictive model should not be influenced by the student's sensitive attributes~\cite{hu2020towards}. Thus, we seek to evaluate to what extent the models are fair regarding these attributes in an Asian context. We train binary classifiers to predict whether a student will receive a grade better than the average and evaluate disparities across fine-grained groups using fairness metrics (see \Cref{subsec:bias_measurement}). 

Lee and Kizilcec~\cite{lee2020evaluation} argued that student success prediction models may introduce bias by failing to identify a successful student ``because of their membership in a certain demographic group''~\cite{lee2020evaluation}. The study focused on the US context and unfairness with respect to race or gender. It evaluated a Random Forest model that predicted whether students from a US university will receive at least a median grade in a course. Based on two fairness measures (equality of opportunity and demographic parity) out of three measures considered, the model exhibited bias against racial minority students and male students.

Jiang and Pardos~\cite{Jiang2021} implemented a recurrent neural network for predicting course grades of students at a US university. Then, they compared the overall model's performance (measured by accuracy, sensitivity, and specificity) against the performance of models when the students were divided into seven groups by race. Since the results exhibited some bias, the authors compared four fairness strategies, out of which adversarial learning achieved the best fairness scores.

Bayer et al.~\cite{Bayer2021} studied a large population of university students in the UK. The performance of models for predicting student success was measured by AUC and false positive/negative rate. For all divisions of student groups (based on ethnicity, gender, or disability status), the models achieved slightly better predictions for the majority group.

A study of three countries -- the US, Costa Rica, and the Philippines -- confirmed the intuition that predictive models trained on data of students from one country generalize better to other students from the same country~\cite{ogan2015towards}. Thus, such usage leads to less bias compared to when the models are deployed in another country. Although this research also examined the Filipino context, it focused on model generalization and not group fairness analysis as in our paper.

\section{Research Methods}
\label{sec:methods}

Our goal is to build and evaluate classification models that predict students' grades from submissions (such as assignments and quizzes) in university courses.

\subsection{Educational Context of the Study}

Our study analyzes data of a population of 5,986 students, all of whom were enrolled in a large university in the Philippines. The data set does not include any external students from other institutions. All students were at least 18 years old, and they were studying undergraduate or graduate programs across different academic disciplines taught at the university in 2021--2022.

Student data were collected from courses that used the university's Canvas LMS~\cite{canvas}. The courses spanned a wide range of offerings: from humanities to natural sciences and engineering. Many of the courses were part of the university's common core curriculum, which includes languages, philosophy, psychology, etc., while other courses were specific to the college majors, e.g., programming or marketing. The course data were combined to ensure broader generalizability (and, if courses were considered separately, the data pool for some of them would be too small). All courses ran fully remote due to COVID-19 restrictions at the time of data collection.

\subsection{Collection of Student Data}
\label{subsec:data}

The research data contain two types of information: (1)~students' usage of Canvas LMS (i.e., their actions in the system) to perform the prediction modeling, and (2) students' approximate location when working with the LMS (i.e., their region) to evaluate the algorithmic bias.

\subsubsection{Canvas LMS Data (Learning Traces)}

\Cref{appendix:data-collection} details the technical aspects of the data collection. Overall, the data of the 5,986 students represent the following possible actions within the LMS:
\begin{itemize}
    \item accessing a learning resource (e.g., viewing a PDF file with lecture slides),
    \item uploading/editing a file (e.g., a homework attachment),
    \item posting in the discussion forum,
    \item submitting an assignment/quiz,
    \item receiving a grade (e.g., for submitting an assignment).
\end{itemize}

Each student in the data set is uniquely identified by a 5-digit Canvas student ID (``student ID'' further in the text). For privacy, this ID is different from the student ID issued by the university. Each course is also identified by a 5-digit ID.

\subsubsection{Demographic Data (Geographic Location)}

For each student ID, our records contain the city and region where the student resided when connecting to Canvas. There are 18 possible categorical values: either one of the 17 official regions defined by the Philippine Statistic Authority~\cite{psa}, or, in very rare cases, abroad (outside the country's borders).

Infrequently, it happened that a student connected to Canvas from multiple regions throughout the three semesters (this could happen when a student visited a family member, for example). In this case, we assigned the student to their majority location, i.e., the one from which they connected the most often throughout the considered time period.

Since some of the 18 regions had very low representations, we subsequently grouped regions that were related, resulting in five larger meaningful clusters. The clusters were determined according to the standard division~\cite{psa} based on the main island groups of the Philippines. The clusters are:
\begin{itemize}
    \item National Capital Region (NCR), i.e., the Metro Manila urban area including the capital city -- 4,816 students,
    \item Luzon (except the NCR) -- 580 students,
    \item Mindanao -- 138 students,
    \item Visayas -- 131 students,
    \item Abroad (see the explanation above) -- 49 students.
\end{itemize}

For all clusters (except Abroad), the locations inside that cluster have mostly similar context in terms of economy, culture, and language. Internet access might sometimes vary even within the regions on a smaller level of granularity. NCR is separated since it is the densest region, and its context is different from the other regions as it is very highly urbanized.

The student counts across the five regions sum to 5,714, since we lacked the location data for 272 students. So, we used the full data set of 5,986 students for the modeling (since these were valid log data from the population), and the data of a subset of 5,714 students for the bias evaluation.

\subsubsection{Data Anonymization and Ethical Issues}

The university had permission to use students' Canvas and location data (which the students were supposed to report) for research purposes. These data are paired only with the pseudonymous 5-digit Canvas student ID, so the researchers handling the data could not access other information about a particular student. All data were anonymized and used in accordance with the country's and university's regulations.

\subsection{Preprocessing of Canvas LMS Data}

Due to the huge size of the raw Canvas LMS data set, we automatically preprocessed it using a dedicated Python script (see \Cref{appendix:materials}). The preprocessing involved computing the values of predictor and target variables, which are explained below. Then, the preprocessed data were used for building models that predict the student grades (see \Cref{subsec:modeling}).

\subsubsection{Target Variable (Grade)}

Since different courses/assignments/quizzes in the data set use different grading scales, we normalized the grade as a decimal number between 0.0 and 1.0, which is the ratio between (a) the number of points that the student received and (b) the maximal number of points that was possible to be obtained in that particular context.

The average grade in the data set is 0.721. For the binary classification task, we aim to determine whether the student will receive a grade \textit{worse than an average grade} (class 1, ``unsuccessful'') or not (class 0, ``successful''). Out of the total of 211,945 grade entries (rows in the preprocessed data), 76,556 (36.1\%) are below the average grade.

\subsubsection{Predictor Variables (Features)}

To predict the grade, we extracted features from events that a particular student performed in Canvas prior to receiving that grade within the context of a particular course. In other words, we explored whether and to what extent are student actions in Canvas predictive of their grade. We used 27 features grouped in the following categories:
\begin{itemize}
    \item Count of events (examples: the number of resources accessed, the number of uploaded homework files, the number of submitted quizzes).
    \item Conditioned count of events (examples: the number of forum posts with a minimal length of 100 characters, the number of submissions that were not past the deadline, the average number of learning resources accessed per day of being active in the system).
    \item Timing of events (examples: the average/standard deviation of the time gap between two accesses of a learning resource or submitting a file).
\end{itemize}

\subsection{Models' Training and Cross-Validation}
\label{subsec:modeling}

To provide a comparison and find a best-performing model, we selected five binary classifiers: Decision Tree (DT), Random Forest (RF), K-Nearest Neighbors (KNN), eXtreme Gradient Boosting (XGB), and Logistic Regression (LR). Their performance was evaluated by AUC ROC and weighted F1-score. As a baseline, we also trained a ``dummy'' classifier, which assigned each data point to the majority class (0), yielding AUC and weighted F1-score of 0.5. We used Python 3.10 and the Scikit-learn library~\cite{Pedregosa2011} for the implementation.

The training set consisted of 211,945 samples and 27 features for all 5,986 students. For each model, feature scaling was performed for each predictor variable to unitize the values used for modeling. In addition, we experimented with feature selection by dropping features one-by-one (backward feature selection) or features that were correlated, but removing any feature worsened the model performance.

Model training involved 10-fold student-level cross-validation and a grid search of hyperparameters only on the training data. The training-validation split was stratified, and all samples for a given student appeared ``exactly once in the test set across all folds''~\cite{scikit-stratifiedgroupkfold}. The evaluation metrics were computed as an average across the 10 cross-validation runs.

\subsection{Models' Bias Measurement}
\label{subsec:bias_measurement}

\begin{figure}[htb]
    \centering
    \includegraphics[width=0.95\columnwidth]{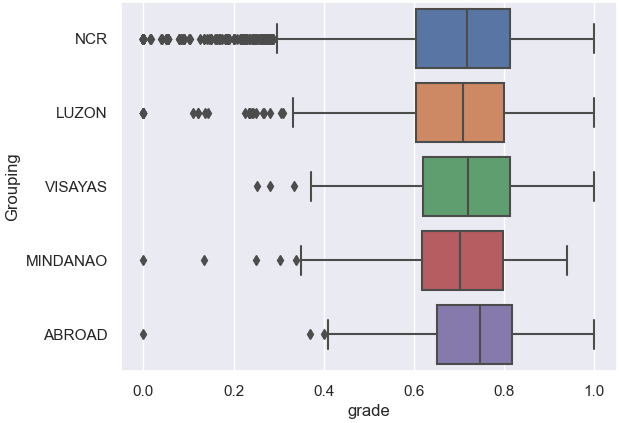}
    \caption{Grade statistics across the five student groups.}
    \Description{Boxplot diagram showing similar grades across all five groups.}
    \label{fig:grade-boxplot}
\end{figure}

As mentioned in \Cref{sec_intro}, we used a group fairness approach to measure algorithmic bias. For an initial check, we examined the distribution of grades in each group to see whether bias or a major difference is already present. The boxplot (\Cref{fig:grade-boxplot}) and kernel density estimate plots~\cite{KDEplot} of grades indicated a similar distribution of grades across groups.

Subsequently, we compared the models’ performance across the five region groups to identify potential unfair predictive results. To do so, we chose complementary metrics: two traditional and one novel. Firstly, we used the two predictive performance metrics used for model evaluation, AUC ROC and weighted F1-score (both ranging from 0 to 1). Secondly, we used a dedicated fairness metric that evaluates the model behavior rather than its predictive performance, MADD (ranging from 0 to 2, with smaller values being better)~\cite{vergerEDM2023}. It evaluates how different the predicted probability distributions are between groups, even though the groups are not provided as features to the model and thus do not influence the probability distributions. 

Thus, the first two metrics evaluate to what extent a model was as accurate for a group as for the others. The third one quantifies to what extent a model produced varying prediction distributions for a group over the others. We computed AUC ROC and weighted F1-score for each group, and MADD for each group against all other students (one-versus-rest approach). Therefore, for a model to be considered fair in this context, it should obtain for each group: AUC ROC and weighted F1-score with low variance (close to other groups), and low MADD score (close to 0).

\section{Results and Discussion}

We first look at the classifiers alone, then evaluate their bias.

\newcolumntype{i}{>{\columncolor{gray!10}}c}
\newcolumntype{j}{>{\columncolor{gray!20}}c}

\begin{table*}[t]
\centering
\caption{Results of modeling and bias evaluation. Average values across the 10-fold cross-validation are listed, with standard deviations in parentheses. The column ``All'' reports the performance of binary classification models on the data of 5,986 students, sorted descending by the AUC value. The remaining columns report per-group model performance on the data of 5,714 students.}

\begin{tabular}{cc|ciiiiijj}
    &      & All             & NCR             & Luzon           & Mindanao        & Visayas         & Abroad          & Mean & $\triangle (\text{All})$ \\ \hline
\multirow{3}{*}{RF}
    & AUC  & 0.75 \g{(0.00)} & 0.75 \g{(0.01)} & 0.74 \g{(0.02)} & 0.76 \g{(0.04)} & 0.75 \g{(0.03)} & 0.77 \g{(0.04)} & 0.75 & 0  \\
    & F1   & 0.79 \g{(0.00)} & 0.79 \g{(0.01)} & 0.78 \g{(0.02)} & 0.80 \g{(0.05)} & 0.78 \g{(0.03)} & 0.82 \g{(0.03)} & 0.79 & 0  \\
    & MADD & \g{N/A}         & 0.14 \g{(0.02)} & 0.17 \g{(0.03)} & 0.31 \g{(0.08)} & 0.31 \g{(0.08)} & 0.55 \g{(0.22)} & 0.30 & \g{N/A} \\ \hline
\multirow{3}{*}{XGB}
    & AUC  & 0.73 \g{(0.02)} & 0.73 \g{(0.02)} & 0.73 \g{(0.01)} & 0.74 \g{(0.03)} & 0.73 \g{(0.03)} & 0.74 \g{(0.05)} & 0.73 & 0       \\
    & F1   & 0.78 \g{(0.01)} & 0.78 \g{(0.02)} & 0.77 \g{(0.02)} & 0.79 \g{(0.05)} & 0.76 \g{(0.04)} & 0.80 \g{(0.04)} & 0.78 & 0  \\
    & MADD & \g{N/A}         & 0.12 \g{(0.02)} & 0.16 \g{(0.03)} & 0.28 \g{(0.06)} & 0.28 \g{(0.09)} & 0.52 \g{(0.20)} & 0.27 & \g{N/A} \\ \hline
\multirow{3}{*}{DT}
    & AUC  & 0.70 \g{(0.01)} & 0.70 \g{(0.01)} & 0.70 \g{(0.02)} & 0.71 \g{(0.04)} & 0.71 \g{(0.02)} & 0.73 \g{(0.03)} & 0.71 & + 0.01  \\
    & F1   & 0.74 \g{(0.01)} & 0.74 \g{(0.01)} & 0.73 \g{(0.02)} & 0.76 \g{(0.05)} & 0.74 \g{(0.02)} & 0.79 \g{(0.04)} & 0.75 & + 0.01  \\
    & MADD & \g{N/A}         & 0.16 \g{(0.04)} & 0.20 \g{(0.06)} & 0.39 \g{(0.08)} & 0.36 \g{(0.05)} & 0.62 \g{(0.25)} & 0.34 & \g{N/A} \\ \hline
\multirow{3}{*}{LR}
    & AUC  & 0.62 \g{(0.00)} & 0.62 \g{(0.00)} & 0.61 \g{(0.01)} & 0.63 \g{(0.04)} & 0.61 \g{(0.02)} & 0.63 \g{(0.04)} & 0.62 & 0       \\
    & F1   & 0.60 \g{(0.00)} & 0.60 \g{(0.01)} & 0.60 \g{(0.01)} & 0.60 \g{(0.03)} & 0.60 \g{(0.02)} & 0.62 \g{(0.03)} & 0.60 & 0       \\
    & MADD & \g{N/A}         & 0.11 \g{(0.04)} & 0.14 \g{(0.07)} & 0.24 \g{(0.06)} & 0.21 \g{(0.05)} & 0.45 \g{(0.19)} & 0.23 & \g{N/A} \\ \hline
\multirow{3}{*}{KNN} 
    & AUC  & 0.55 \g{(0.01)} & 0.55 \g{(0.01)} & 0.55 \g{(0.01)} & 0.55 \g{(0.02)} & 0.54 \g{(0.03)} & 0.58 \g{(0.09)} & 0.55 & 0       \\
    & F1   & 0.59 \g{(0.01)} & 0.59 \g{(0.01)} & 0.59 \g{(0.01)} & 0.62 \g{(0.06)} & 0.57 \g{(0.04)} & 0.62 \g{(0.10)} & 0.60 & + 0.01  \\
    & MADD & \g{N/A}         & 0.15 \g{(0.05)} & 0.19 \g{(0.07)} & 0.28 \g{(0.07)} & 0.30 \g{(0.05)} & 0.47 \g{(0.20)} & 0.28 & \g{N/A} \\
\end{tabular}
\label{tab_fairness_results}
\end{table*}

\subsection{Binary Classification Models}

In \Cref{tab_fairness_results}, the column ``All'' reports the prediction results, with standard deviation reported for each measurement in parentheses. Although the performance of Logistic Regression and KNN is rather unsatisfactory, they are still better than the naive baseline, and the other three models perform well enough to warrant the bias evaluation.

Within the literature on student performance prediction modeling, our best obtained AUC of 0.75 is slightly on the lower end compared to what was reported in the past. Examples of related work reported AUCs from 0.83 to 0.91~\cite{Rohani2023early} or up to 0.95 depending on the week of the semester~\cite{Jang2022}. Nevertheless, 0.71~\cite{Asselman2023}, 0.75~\cite{sahlaoui2023empirical}, and 0.77~\cite{alamgir2023} (as the best-performing among the compared models) were recently published.

\subsection{Algorithmic Bias}

\Cref{tab_fairness_results} reports the bias evaluation. Each model produced consistent, even constant, AUC and weighted F1-score for each of the region groups. For instance, for the Random Forest model, the AUC ranged from 0.74 to 0.77 and the weighted F1-score from 0.78 to 0.82 for all the groups. Thus, the models did not produce discrepancies in predictive performance between groups.

The MADD results also indicate that the models are quite fair, as they are much closer to 0 than to 2. For instance, for the Random Forest model, per-group MADD ranges from 0.14 to 0.55. For all groups other than Abroad, MADD is on the low side with low variance, meaning that the models generate similar probability distributions across the groups. The highest MADD values are always for the Abroad students, likely because this group is the least represented. Thus, there might not be enough predictions for Abroad students to compare their probability distributions against other groups.

Overall, we did not observe evidence of unfair predictions -- even despite the fact that three groups (Mindanao, Visayas, and especially Abroad) had smaller numbers of students. Smaller samples have more variance and therefore might exhibit more bias, and although there was some indication of this for the Abroad group, it was not substantial. 

Since the average grade (0.72) and the median grade (0.83) in the data set were relatively high, one could argue that teachers' grading was more lenient during the mandated online learning, as reported in~\cite{moreno2023cura}. This skew of the grades, along with the similar grade distribution between the groups (see \Cref{fig:grade-boxplot}), may explain the low variance across the groups. However, we obtained enough differentiation to build accurate models, so this limitation does not seem to apply.

These findings represent a contrast to the previous literature on bias in education; for example, the regional distribution of students mattered in~\cite{Ocumpaugh2014}, which discovered failures to generalize across urban, suburban, and rural populations across regions of two states in the US. The performance of predictive models trained on separate populations significantly differed from each other and also from the overall model. However, this effect pertained to a younger population (K-12 students rather than university students) and for a different prediction problem (affect detection rather than grades).

\section{Conclusion and Future Work}

This paper assessed fairness of prediction models for grades in the context of online learning during COVID-19 lockdowns. The models were evaluated in a novel setting: for multiple groups located across the Philippines, with groups having very different numbers of students. Contrary to our expectations, the evaluation did not demonstrate the consequences of the digital divide in the Philippines. Machine learning models for predicting student grades worked comparably well for students all across the country (and abroad). Supplementary materials for this work are linked in \Cref{appendix:materials}.

While this is a promising result, further research is needed to demonstrate the generalizability of these findings in different contexts in the Philippines and beyond. While the location did not influence the results here, it mattered in other past research for a different prediction task~\cite{Ocumpaugh2014}, so the EDM field needs to further investigate where it does and does not matter. Moreover, the smaller samples for some groups might not be entirely representative of the respective regions, hence, there is a need to replicate the results on larger samples.

Future work should also research whether bias manifests not only in groups defined along a single dimension (e.g., the location), but also across intersections of demographic variables~\cite{Li2023review, Xu2024} (e.g., the student's location and age). This deeper investigation can identify specific biases that might get lost or evened out in an overall evaluation.

\section{Acknowledgments}

We thank Ezekiel Adriel D. Lagmay for data cleaning and anonymization. Valdemar Švábenský thanks Ateneo de Manila University for providing the LS Internationalization Grant that enabled him to visit and collaborate with the research group of Maria Mercedes T. Rodrigo. Mélina Verger's work was supported by Sorbonne Center for Artificial Intelligence (SCAI) and the \textit{Direction du Numérique Educatif}.

\bibliographystyle{abbrv}
\bibliography{references}


\newpage
\appendix

\section{Technical Details of the Data Collection from Canvas LMS}
\label{appendix:data-collection}

The data collection took place over the period of three consecutive semesters in two academic years (AY):
\begin{itemize}
    \item Second semester of AY 2020--2021, divided into two quarters, from February 10, 2021 (with the logging starting on March 9, 2021) to April 7, 2021, and then from April 12, 2021 to June 5, 2021.
    \item First semester of AY 2021--2022, which ran from August~26, 2021 to December 18, 2021.
    \item Second semester of AY 2021--2022, which ran from January 31, 2022 to May 28, 2022.
\end{itemize}

\begin{figure}[htb]
    \centering
    \includegraphics[width=\columnwidth]{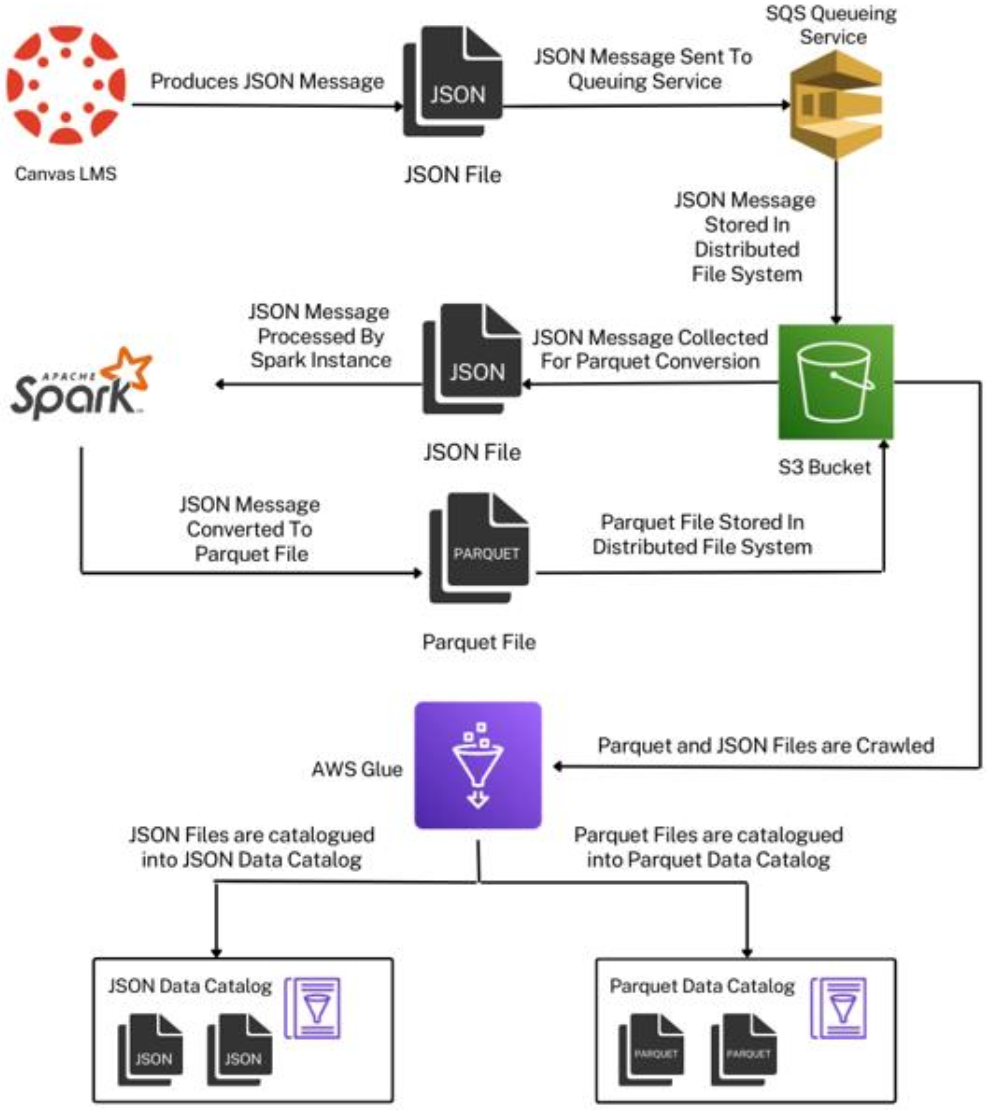}
    \caption{Architecture of the data collection from Canvas.}
    \Description{Diagram of data flow and preprocessing steps.}
    \label{fig:canvas-data-collection}
\end{figure}

\Cref{fig:canvas-data-collection} illustrates the data collection architecture for this project. Events from Canvas in JSON format were sent to a message queue for preprocessing and anonymization using open-source software created at the university (\url{https://github.com/zzenonn/canvas-data-collector}). 

The data then underwent an Extract, Transform, and Load (ETL) process using Amazon Web Services tools. Finally, the data were converted to the format Parquet~\cite{Parquet} for faster analysis and efficient storage.

For this research, the raw data were subsequently exported as a CSV file. The size of this file is more than 7.6 GB, and it contains 15 columns and 48,740,270 rows (i.e., 8,142 entries on average per student), describing numerous activities and learning behaviors throughout the three semesters. Each action (one row in the CSV file) contains a timestamp with millisecond precision.

\section{Supplementary Materials}
\label{appendix:materials}

To support reproducibility within the EDM conference~\cite{Haim2023edm}, our code is publicly available at:

\url{https://github.com/pcla-code/2024-edm-bias}.

We are unfortunately unable to share the data due to the regulations of the university in the Philippines.

\end{document}